# Physical Modeling Techniques in Active Contours for Image Segmentation

Hongyu Lu,    Shanglian Bao

Beijing City Key Laboratory of Medical Physics and Engineering, Peking University, China

**Abstract:** Physical modeling method, represented by simulation and visualization of the principles in physics, is introduced in the shape extraction of the active contours. The objectives of adopting this concept are to address the several major difficulties in the application of Active Contours. Primarily, a technique is developed to realize the topological changes of Parametric Active Contours (Snakes). The key strategy is to imitate the process of a balloon expanding and filling in a closed space with several objects. After removing the touched balloon surfaces, the objects can be identified by surrounded remaining balloon surfaces. A burned region swept by Snakes is utilized to trace the contour and to give a criterion for stopping the movement of Snake curve. When the Snakes terminates evolution totally, through ignoring this criterion, it can form a connected area by evolving the Snakes again and continuing the region burning. The contours extracted from the boundaries of the burned area can represent the child snake of each object respectively. Secondly, a novel scheme is designed to solve the problems of leakage of the contour from the large gaps, and the segmentation error in Geometric Active Contours (GAC). It divides the segmentation procedure into two processing stages. By simulating the wave propagating in the isotropic substance at the final stage, it can significantly enhance the effect of image force in GAC based on Level Set and give the satisfied solutions to the two problems. Thirdly, to support the physical models for active contours above, we introduce a general image force field created on a template plane over the image plane. This force is more adaptable to noisy images with complicated geometric shapes. The experiments have demonstrated that the proposed models induce a highly valuable framework for Active Contours.

## 1. Introduction

Active contours or Deformable Models [1] includes the Parametric Active Contours based on elasticity theory which is usually named as Snakes [2], and Geometric Active Contours (GAC) [3-4] based on Level Set method. Snakes exhibits its advantage in shape extraction from the blurred images with high noises and large gaps while the GAC is preferable for multiple shapes recovery.

Due to method of solving the Snake differential equations directly, the implementation of topological adaptability of Snakes is still complicated in practice, especially for 3-D segmentation. The T-Snake model [5] is the pioneer of this attempt. In this paper, we present a visual dynamic model, whose validity can be proved based on several independent procedures. It can be repeated with less effort and be adopted in the general framework of Snakes. The concept of region burning is introduced to transfer the topological flexibility problem to a question of simulating the movement of the fire front on the prairie. The key point is to realize the topological change at the final stage of the segmentation process, while in the traditional approach such transformation is done whenever the collision of two curves is detected. By simulating the balloon inflation in a closed space and treating the parent Snake contour as a firewall and the swept area as a burned region, the child Snakes contours corresponding to each object can be extracted more effectively.

. The classical GAC model suffers from the large gap problem which could be partly solved by adding a penalty term including area information [6]. However, so far it is still not clear how to add the image force to the Level Set formulation in low order numerical scheme. The Mumford-Shah energy formulation of GAC like C-V model [7] can give the exact outline of the shape boundaries. This type of model is sensitive to the weighting parameters, and is difficult to process the non-homogeneous images with a large gap in the edge map. By using the physical modeling techniques, we propose a novel method to separate the segmentation procedure into two steps. The idea is to treat the curve evolution process at the final stage as the wave moving in the isotropic material, instead of the anisotropic material in which the speed and direction of the curve controlled by image force field in the GAC equation in [4]. The first step does not include the image force, the propagation term driving the contour evolution. In the second step, the image force directs the evolvement of the contours. If there is a cross-border initialization of the contour, it needs an extra step by employing the image force to attract the contour to the border before taking the first step.

The image force is an essential factor to guide the evolvement of Active Contours, and to provide the critical information to maintain high accuracy of the segmentation. The most influential models are gradient map, Distance Map [8], GVF [9], and the force model coming from nature physics forces such as gravity and electric field [12]. In this paper, we extend the works in [10-11] on the basis of electric field to a general image force model, by which the movement of the contour can be controlled more effectively. Unlike CPM [12] model, this force is not required to follow the Coulomb law. It is computed on a template layer over the image plane to prevent the contours from being trapped by noise points or spurious edges.

The whole paper is organized as follows: A discussion of the feature necessary for constructing an effective image force field is given in Section 2, where the structure and formulation of our generalized image force model are also introduced and analyzed. In Section 3, we illustrate the technique for multiple shapes extraction with the simulation of the inflating balloon and region burning, which is the main contribution of this paper. The techniques to tackle the problems of large gap and segmentation error in GAC are described in Section 4. The experimental results to demonstrate the proposed modeling method are given in Section 5.

## 2   A Generalized Image Force Model

### 2.1   Background

Normally, there are two forces in the active contours. One is the internal force coming from the definition of active contours itself. The other is the external forces such as the image force dependent on the features of image, and the force dependent on contour geometry, for example, the balloon force [13] in Snakes. Image force is the essential factor to ensure the accuracy and performance of the segmentation process.

From the perspective of curve evolution, normally, the image force acquired by edge maps directs the active contours to extract the object boundaries. The image force based on region information can be formed in the similar way as an additional regulator. Hence, our study focuses on the formulation of image force field based on edge map. Based on theory of the thermodynamics and fluid mechanics, the GVF model can construct an image force field with long capture range and maintain the segmentation accuracy. The models such as GGVF [14], NGVD [15] were then developed to push the contour into significant concave or convex, or to alleviate the leakage problem [16]. Another approach utilizes the Newton's gravity law or Coulomb's law to suit the rule for the image force discussed in 2.2. This approach is reflected works in VEF [17] and SSEF [18] which show that application of electric field as image force has similar performance to GVF.

As discussed in [10-11], the GGVF and NGVD models are problematic in the T-junction. This can be illustrated in the Figure 1. The dash arrow in Figure 1 (a) and (b) represents the direction of the image force field generated by GGVF or NGVD models. The solid arrow indicates the direction of the movement of the Snakes. The initial contour is a large circle outside the convex in Figure 1 (a) or a small circle in the T-shape in Figure 1 (b). In Figure 1 (a), under the impact of image force, the circle can shrink into the convex because the image force points to the same direction as the movement of the Snake contours. In Figure 1 (b), the direction of the image force field prevents the initial Snake circle going upward. In the medical images like blood vessel, there are lots of the tubular shapes with many T-shape branches. It is clear that the GGVF and NGVD models cannot deal with such kind of images.

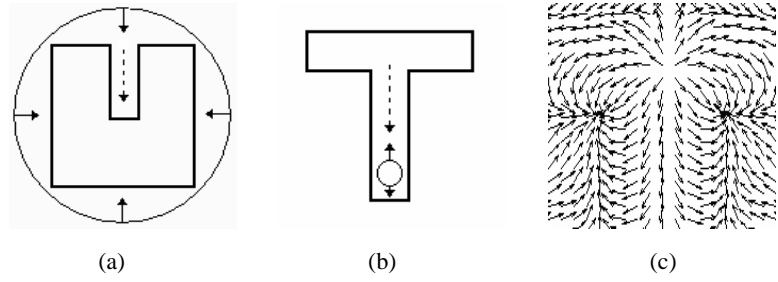

(a)　　　　　　　(b)　　　　　　　(c)

**Fig. 1.** Problem of GGVF and NGVD models at T-junction. (a) For the convex, both models work well. (b) For the T-junction, both Model cannot work. (c) The distribution of the vector field of the image force at T junction of NGVD.

### 2.2　Features of the Image force

From many tests and the analysis above, in most cases, the features of image force should comply with the following rules

1) The ability to maintain the segmentation accuracy should be guaranteed in the force model. Furthermore, the image force should help the active contours to indicate a reasonable boundary of a blurred object.

2) The image force field allows the contours moving easily in tubular shapes, especially at the T-junction.

3) The capture range of the image force can reach far enough to spread the influence of the salient edges to the large edge gaps.

4) The distribution of the image force field should be predictable to facilitate the adjustment of the parameters. If the formula of image force conforms to the common sense, for instance, expressed with simple physical laws, it is not difficult to achieve this goal.

5) The image filed should encourage the ability to free the contours trapped by spurious edges and noise points. This could be achieved by improving the influence of the salient edges in the construction of the image force field.

6) The computation speed should be fast enough in order to process huge datasets.

### 2.3　The Image Force Model

Following the rules in section 2.2, we generalize the image force model in [10] to a practically adaptive formulation. This force model specifies a 2-D image force field which is calculated on a template plane directly over the image plane as the structure shown in Figure 2.

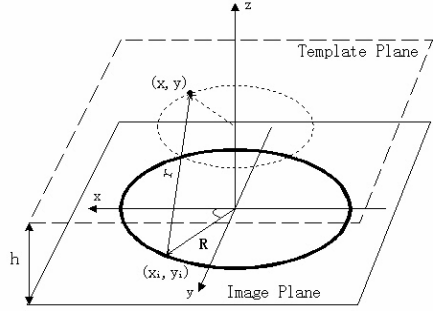

**Fig. 2.** Structure of the image force field

The gray level $g_{x,y}$ of the pixel in the position $(x_i, y_i)$ on the image plane contributes a potential on the template plane. The total potential in position $(x, y)$ on the template plane created by the edge map like the black circle on the image plane can be calculated from the formula below

$$\varphi(x, y) = \sum_{y_i}\left[\sum_{x_i} g_{x_i, y_i} \cdot f(r)\right] \qquad (1)$$

where $g_{x_i, y_i}$ represents the gray level of point at $(x_i, y_i)$ on the image plane, $r = \sqrt{(x-x_i)^2 + (y-y_i)^2 + h^2}$, and $|x-x_i| \leq x_{max}$, $|y-y_i| \leq y_{max}$, or $r < r_{max}$. The $f(r)$ is a transfer function responsible for creation of the image force, which can be in various forms, typically can be written as

(a) $f(r) = 1/(k \cdot r^p)$,

(b) $f(r) = c^{-z}$, where $z = k \cdot r^p$. The most useful expression is: $f(r) = e^{-kr}$.

(c) $f(r) = \pi/2 - \arctan(k \cdot r^p)$

where $k$ and $p$ are both positive real numbers.

A group of variations to the formats above can all be easily created. In computing $r$, the most important parameter which has to be fixed is $h$, the distance between two planes. The black circle in Figure 2 is used to estimate the value range of $h$. It can be seen in Figure 3 that the error of the potential field is kept within 1 pixel if $h \leq 2$. From our tests, the distance $h$ should be a positive real number in the range of $(0, 2]$. In contrast to the discrete image pixels, setting of $h$ to integers will generate larger errors as $h$ increases. In addition, it was described a fast approach for calculation of the potential field in [11].

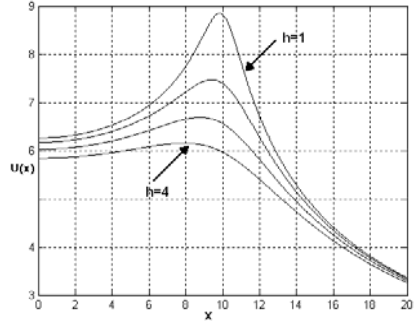

**Fig. 3.** The potential distribution $U(x)$ of the circle with positive charges on the template plane, $R=10$, $h=1, 2, 3, 4$, $p=0.5$ with using format (a) in equation (1)

The features of generalized models are the similar to the basic model in [10-11]. The capability of changing the distribution of the force field by regulating $h$ to minimize the impacts from noise and spurious edges is discussed in [10]. From equation (1), the image force at position $(x, y)$ on the template plane is obtained as

$$F_x(x, y) = \frac{\partial \varphi(x, y)}{\partial x} \qquad (2)$$

$$F_y(x, y) = \frac{\partial \varphi(x, y)}{\partial y} \qquad (3)$$

## 3   Multiple Shapes Recovery by Snakes

### 3.1   Background

Compared with the Geometric Active Contours, Snakes itself can not split the original contour to child Snakes or to merge several contours to one contour automatically. This problem comes from the fact that the Deformable Models is developed on the basis of elasticity theory. This difficulty comes from the fact that the evolvement of Snake contour is a process of solving simply two differential equations. Therefore, this question is better settle with the technique in graphics. The current study on multiple shapes extraction focuses on detecting the collision of two Snake curves. The key of this approach is that once two pieces of Snakes curves encounter, it starts to merge. A technical issue in this method is how to discover the collision of the two curves instantly during the evolvement of Snakes.

Besides an Affine Cell Image Decomposition (ACID) method in T-Snakes, Delingette utilized two topological operators on the basis of grid-based data structure to segment multiple objects in [19]. Loop Snakes adopts the CEP method and the structure of Loop-Tree in the segmentation process [20]. An edge preserving technique was developed based on GVF force to accomplish the recovery of multiple shapes in [21]. Lachaud invented a metric using contour curvature to settle this difficulty in [22].

### 3.2 Topological Changes of Snakes based on Balloon Model

From a different point of view, we can see that it is unnecessary to merge the Snakes curves each time when the collision happed. On the other hand, it is wise to make the topology changes after all the collisions have happened. This idea leads to a physical model, which simulates the process of a balloon expanding in a constraint space.

Suppose there are several small objects within a large closed spherical shell. The shell also serves as an object to be extracted, and the initial Snakes is a balloon surface. The balloon continues to expand until the inflating process stops and all the interspaces are filled by the balloon except for the objects. If removing the self-touched balloon surfaces, the remaining balloon surfaces can be regarded as the child-snake corresponded to each objects.

During the evolvement of the Snake contour, the image force of the equation (1) is adopted. To improve the ability of the Snakes to move in the long tubular shapes, the balloon force needs to be included in the Snakes equation. The smoothing term is omitted because it reduces the accuracy of the segmentation and adds the load of computing. Therefore, the equations for the Snakes in 2-D domain are established as

$$\frac{\partial x}{\partial t} = \alpha \frac{\partial^2 x}{\partial s^2} + \gamma F_x(x, y) + \lambda n_x(x, y) \qquad (4)$$

$$\frac{\partial y}{\partial t} = \alpha \frac{\partial^2 y}{\partial s^2} + \gamma F_y(x, y) + \lambda n_y(x, y) \qquad (5)$$

where $n_x(x, y)$ and $n_y(x, y)$ are the balloon force terms if the points on initial Snakes contour is initialized clockwise. They can expressed as

$$n_x = dy / \sqrt{(dx)^2 + (dy)^2} \qquad (6)$$

$$n_y = -dx / \sqrt{(dx)^2 + (dy)^2} \qquad (7)$$

### 3.3 Implementation with Two Segmentation Stages Using Balloon Model

The critical issue of multiple shape recovery is to remove the self-touched surfaces in 3-D domain or curves in 2-D domain. As discussed above, the solution is to trigger topological changes after the Snakes contour stops evolving. To meet the entropy condition describe in [5], the region swept by the Snakes is marked as a burned region. The border of the burned region, which can be treated as the fire front, will follow the evolvement of Snakes contour with a distance of 1-2 pixels. Thus, with the help of the burned region, the Snake contour which functions as a firewall, can carry a satisfied solution to realize the balloon model. There are two segmentation stages for implementing the Balloon model. The first stage is to stop the movement of the Snakes contour according to a certain criterion. The criterion is set as: the evolvement of Snake curve terminates when it collides with the burned region. The second stage is to remove the self-touched surface and form a connected region with holes.

We illustrate topological transformation through segmenting an artificial image containing two objects in an image with the small circle and a large circle. The initial Snake contour (in green color) is at the bottom left inside the large circle as show in Figure 4 (c). The movement of the Snake contour is controlled by equations (4) and (5) as illustrated in Figure 4 (b). The whole process is depicted in Figure 5 and Figure 6 with emphasis on how to remove the touched contours in 2-D space domain. For clarity, the distance from fire front (in red color) of the burned region (in blue color) to the Snakes contours (in green color) is set to 2-3 pixels.

Since the burned region is within the area swept by Snakes, the contour will stop moving after two pieces of curve collide into the border of opposite burned region as shown in Figure 4 (c). Then the intersected curves form a narrow band which is an unburned band as shown in Figure 4 (d). This narrow band can be imaged as the touched walls of the balloon surface.

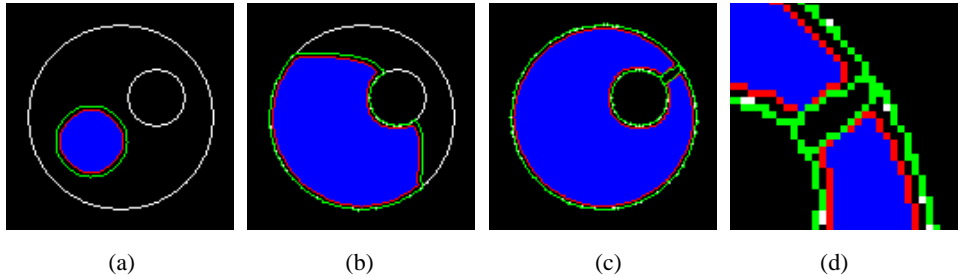

      (a)            (b)           (c)           (d)

**Fig. 4.** Evolvement of the Snakes contour and termination of the movement (a) Initialization of the contour. (b) Movement of the contour the burned region. (c) Intersection of the Snake curves and their collision with the burned region. (d) Zooming out of the intersection of two curves.

When arriving at the stable stage, neither Snakes contour nor the fire front can go forward anymore because the direction of their movement is opposite. At this moment, if we move the Snake contour for a piece for a while again, the burned region can get free from the restriction of the Snake curve. This procedure resembles the action of moving the Snake firewall into the burned region, which is shown in Figure 5 (a). Now the fire front can continue to burn until creating a connected region as show in Figure 5 (b). Extracting the boundaries (in cyan) of the burned region as shown in Figure 5 (c), it can acquire the child Snakes contour corresponding to each object. The contours of the child Snakes need only to evolve several time steps to get the exact boundaries of objects, which are two circles, as the result given in Figure 5 (d).

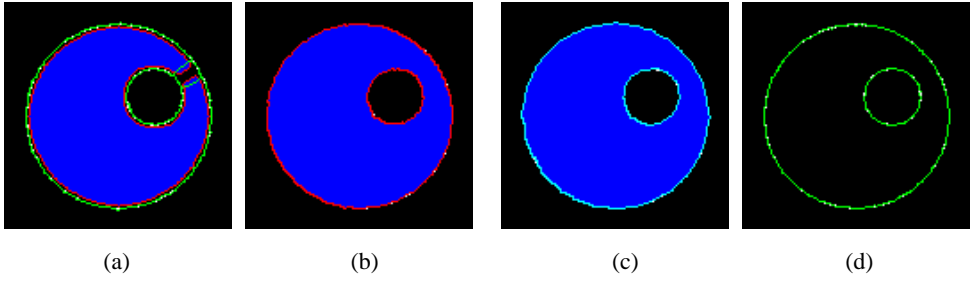

(a)                (b)                (c)                (d)

**Fig. 5.** Recovering boundaries of the small circle and large circle by continuing to evolve the Snake contour, and completing the region burning. (a) Evolving the Snake contour again. (b) A connect region is formed after burning out the narrow band. (c) Extracting the boundaries of the burned region. (d) Final status of two child Snakes after extracting the boundary of the burned region and evolving them for a piece of while.

## 4    Accurate GAC Segmentation of Image with Large Gaps

### 4.1    Background

Compared with the newly developed approach such as C-V model, the classical GAC have several advantages: (a) Segmentation result can be predicted by the parameters more easily. (b) It can get the specified objects as desired. (c) It is robust, stable. (d) It can process object with inhomogeneous gray levels such as the tumor in medical images more confidentially.

Due to the loss of the elastic force in Snakes, the large gap on the edge map always makes the contour leaking from breaks. Furthermore, the detail of incorporating image force into GAC equation (8) has not fully and clearly disclosed. These problems limit the application of GAC in practice. The segmentation error in the pioneering paper [3], which is about 2 pixels, still exits in the programs available online even in the GAC algorithm provided in the ITK [23]. Without a complex high order scheme as [24], image force is difficult to work well in the GAC expression based on Level Set. To explore this problem, a steering function derived from histogram features is created in [25]. In this paper, we take the equation for this GAC as below, which is same as that in [4].

$$\frac{\partial \phi}{\partial t} = -\lambda k_I |\nabla \phi| + \varepsilon \kappa |\nabla \phi| - \gamma F \bullet \nabla \phi \qquad (8)$$

where $\lambda, \varepsilon$, and $\gamma$ are weights for propagation, smoothing, and advection by image force.

### 4.2    Simulation of the propagations of the wave in the isotropic-substance

As we know, the wave propagates in isotropic-substance always at the same speed. It only switches to the inverse direction when encountering an obstacle. This model can be expressed simply with the term $\gamma \text{sign}(F \bullet \nabla \phi)$ in equation (10) below. In contrast, the term $\gamma F \cdot \nabla \phi$ in equation (8) simulates the wave moving in the anisotropic-substance, which changes the speed and direction of the wave front according to the information of the image force field. This model leads to a complex and high order solution. The isotropic model provides a clear way to realize the wave propagation, which can be easily added to the curve evolution process based on Level Set. From this point, a procedure with several segmentation stages has been developed to solve the problems of the large gap, segmentation error and the cross border in classical GAC. This scheme is described as follows

Step 1: Only the Propagation force and the smoothing term working.

$$\frac{\partial \phi}{\partial t} = -\lambda k_I |\nabla \phi| + \varepsilon \kappa |\nabla \phi| \qquad (9)$$

Where the term $k_I = 1/\left(1 + \left|\nabla P_E^h\right|^m\right)$, $P$ means the potential field.

Step 2: After the GAC contour stop movement, the Image force takes actions to simulate the wave transferring in the isotropic-substance

$$\frac{\partial \phi}{\partial t} = \varepsilon \kappa |\nabla \phi| - \gamma \, \text{sign}(F \cdot \nabla \phi) \qquad (10)$$

where the sign of the $F \cdot \nabla \phi$ is decided by

$$\text{sign}(F \cdot \nabla \phi) = \max(F_x, 0) \cdot D_{ij}^{-x} + \min(F_x, 0) \cdot D_{ij}^{+x} + \max(F_y, 0) \cdot D_{ij}^{-y} + \min(F_y, 0) \cdot D_{ij}^{+y} \qquad (11)$$

For the initial contour with the cross border problem, the Step 2 should be executed as an extra step before executing Step 1 to attract the initial contour (in yellow) to the object boundaries (in blue). Figure 6 shows the segmentation result using this scheme to process an initial contour with the cross border problem. Figure 7 gives the estimated outline of the boundary at large gaps with the help of the image force model in Section 2.

Using Fasting Marching method to reconstruct the Signed Distance Function (SDF) can accumulate error gradually. Instead, we adopt the Level Set – Narrow Band method by designing a discrete circle stencil, in which the distance of each point to the central point has been calculated in advance. By moving this stencil on the contour of the zero level set and finding the minimum value of each point during the motion, the SDF in the narrow band can be given precisely. Compared with using the square as the stencil, the circle stencil can remove the large error caused by the four corners of the square in the narrow band because the image is discrete. The experiments prove that this technique can speed the computing process up to tens of times in comparison with the direct calculation method. Specially, we should point out, in the application of image analysis, the SDF can be taken as the format like $k \cdot r$ and $r^2$ to get peculiar segmentation results.

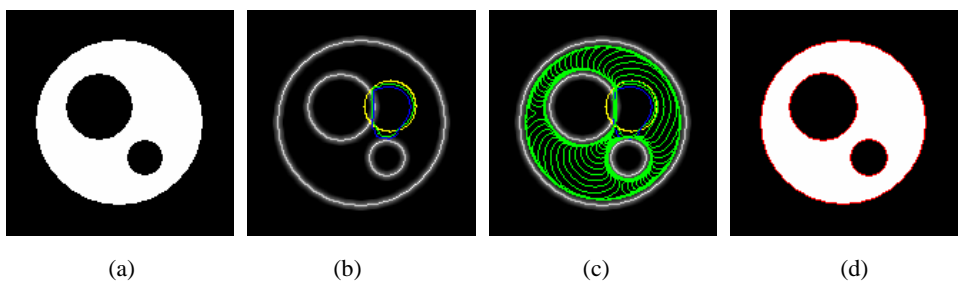

**Fig. 6.** GAC segmentation of an image with three circles.   (a) Original image   (b) The gradient edge map and the initial GAC contours), and the result of the extra segmentation stage for attracting the contour to object boundaries.   (c) The evolving contours (in green) of the fist segmentation stage (Step 1).   (d) The result of the second segmentation stage (Step 2), three circles are extracted.

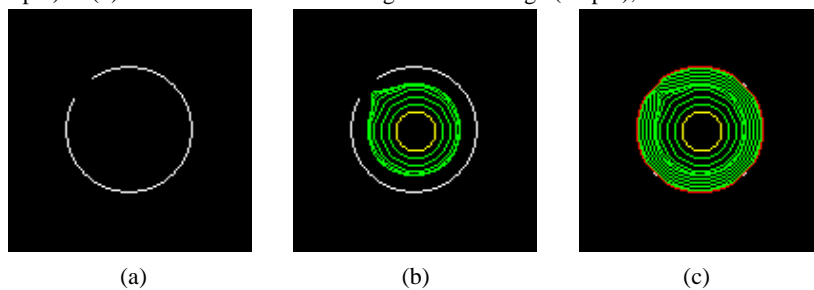

**Fig. 7.** GAC segmentation of an image with three circles.   (a) Original image (b) Initial GAC contours and the process of the evolving contours (in green) of the first segmentation stage (Step 1). (c) The segmentation result (in red) of the second segmentation stage (Step 2), the GAC contour fills the gap on the circle.

## 5   Experiment Results

Two images are tested using the Snaked based on the Balloon model. The equations (4) and (5) are adopted in shape extraction. Figure 8 shows the segmentation of the brain bone. There are totally seven objects with significant edge features. By using the scheme based on the Balloon model, all the shapes are extracted correctly. The segmentation of the cells by Snakes is given in Figure 9. Seven cells with three abnormal nucleoli and three cell fragments are identified correctly. In this case, the Snakes contour is deflating.

Two images are tested using the GAC scheme above. During the different segmentation stage, the equations (9), (10) and (11) are applied respectively. Figure 10 (b) and 11 (b) illustrate the segmentation procedure of the proposed GAC scheme. In Figure 10, the cross border problem is solved and the accuracy of the segmentation is well guaranteed. The segmentation of a CT image with two gaps on the border of the body is given in Figure 11. This image suffers from the truncation problem from the CT scanning. Through setting threshold to the edge maps and applying the GAC algorithm above, two large gaps on the body border are successfully linked by the GAC contour. Then EM algorithm can be utilized to complete the electronic cleansing in the colon for Virtual Colonoscopy [26]. The proposed GAC method can be extended to the 3-D approach. Figure 12 shows the 3-D shape extraction from the MR imaging of a bladder with a tumor.

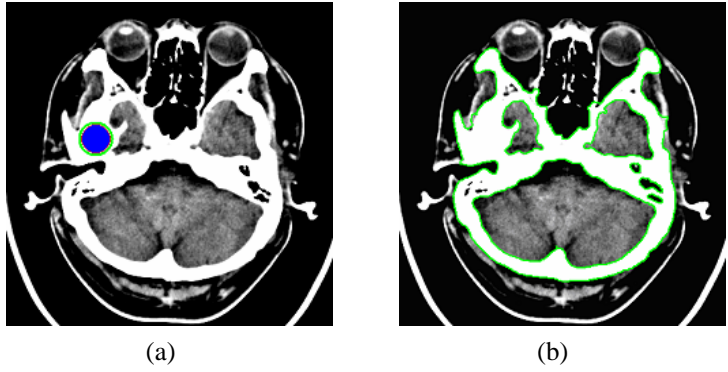

(a)                                                             (b)

**Fig. 8.** Using the Snakes and the Balloon model to segment bone of the brain. (a) Original Image, initial Snakes contour (in green), initial burned region (in blue) and fire front (in red). (b) The segmentation result (in green) with seven objects of brain bone

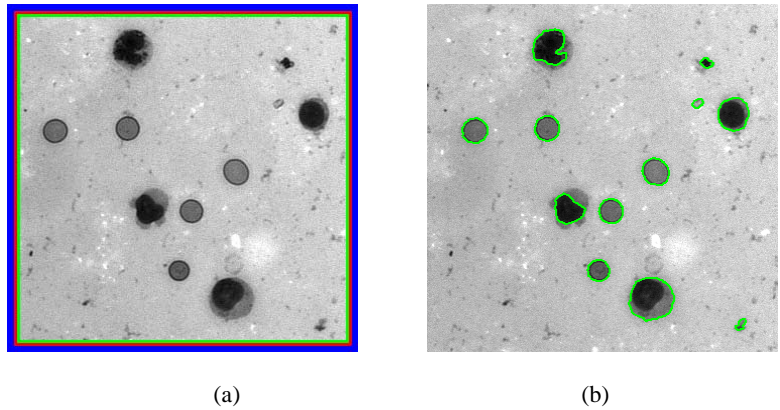

(a)                                                           (b)

**Fig. 9.** Using the Snakes and the Balloon model to extract cells. (a) Original Image, initial Snakes contour (in green), initial burned region (in blue) and fire front (in red). (b) The segmentation result (in green) of nine objects.

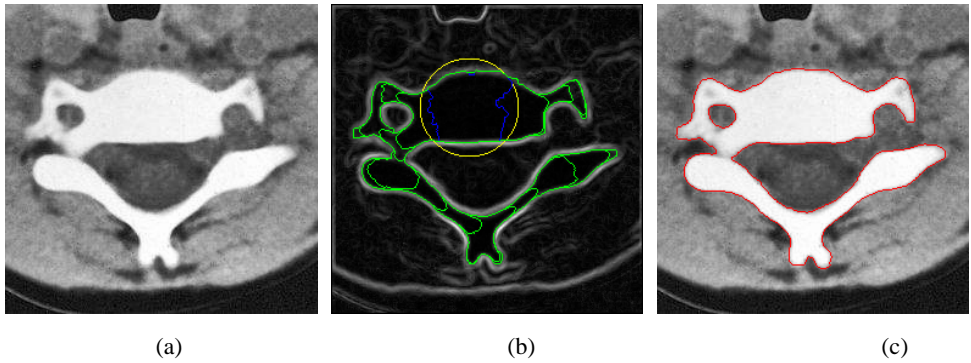

(a)                                 (b)                               (c)

**Fig. 10.** Using the GAC to extract the neck bone. Original Image. (b) Initial Snakes contour (in yellow), the evolvement result (in blue) of Snakes in the extra stage for attracting the contour to the object boundaries, the evolvement of Snakes in the first stage (in green, Step 1). (c) The segmentation result of the second stage (in red, Step 2). The segmentation accuracy is satisfied by using the proposed GAC scheme.

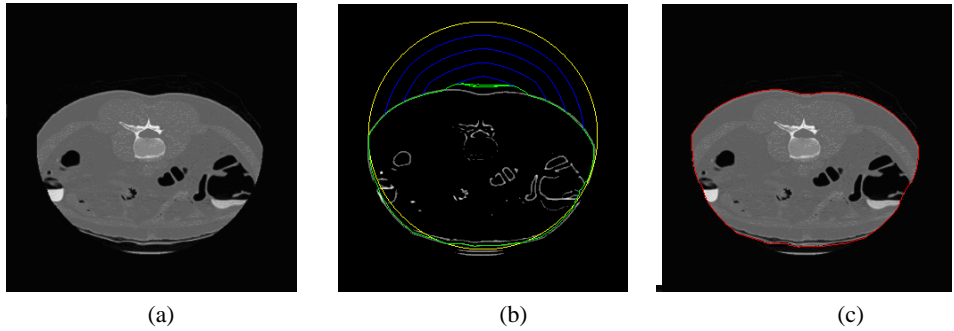

**Fig. 11.** Using the GAC to link the gaps of a CT Image. (a) Original Image. (b) Initial Snakes contour (yellow) and the evolvement result (blue) of Snakes in the extra stage for attracting the contour to the object boundaries, and the evolvement of Snakes in the first stage (green, Step 1). (c) Segmentation result of the second stage (red, Step 2). Two gaps are connected by the GAC contour.

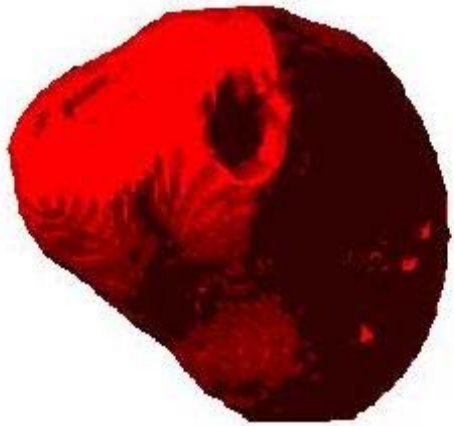

**Fig. 12.** The 3-D segmentation result of a MRI bladder with tumor by GAC.

## 6  Conclusion

In this paper, we follow the concept of physical modeling to develop the techniques for solving the problems in active contour models. Physics principle, visual models and process simulation are induced in the segmentation procedure. The Balloon model can provide a concise solution to multiple shape recovery problems in Snakes. Dividing the simulation process into several stages according to physics principle can solve the large gaps problem, and satisfy the demand for accurate shape extraction for GAC method. In addition, we extend the formulation of the image force model, to support the evolution of the Snakes and GAC contours. From the experiments, we demonstrate that the physical modeling is powerful and is able to deal with the difficulties in active contours. Since the extension of the Balloon model to 3-D segmentation is straightforward, our future work will emphasize on this topic. Investigation of the features in this image model is also a focus of our next research.